\documentclass[conference]{IEEEtran}
\IEEEoverridecommandlockouts

\usepackage{cite}
\usepackage{amsmath,amssymb,amsfonts}
\usepackage{graphicx}
\usepackage{textcomp}
\usepackage{xcolor}
\usepackage{booktabs}
\usepackage{array}
\usepackage{url}
\usepackage{balance}
\usepackage{subcaption}

\title{Safety-Oriented Routing Analysis of Mixtral MoE Under Benign and Harmful Prompts}

\author{
\IEEEauthorblockN{Md Nurul Absar Siddiky }
\IEEEauthorblockA{Department of Electrical and Computer Engineering\\
University of Hawai'i at M\={a}noa\\
Honolulu, HI, USA\\
msiddiky@hawaii.edu}
\thanks{Course: EE609 -- Computer and Network Security. Instructor: Dr. Yingfei Dong.}
}

\begin{document}
\maketitle

\begin{abstract}
Sparse mixture-of-experts (MoE) language models activate only a small subset of parameters for each token, making router behavior a central part of model computation. This paper studies routing behavior of Mixtral 8x7B-Instruct under benign and harmful prompts using two complementary signals: activation-based routing scores derived from expert selection frequencies and gradient-based scores derived from router-gate sensitivities. We analyze expert- and layer-level routing behavior and conduct expert-suppression interventions. The results show that activation-based expert usage is broad and long-tailed, whereas gradient-based importance is concentrated. At expert level, benign and harmful prompt groups remain close under both signals with modest separation. At layer level, activation-based routing is most selective around layers 8-15, while gradient-based importance is concentrated in final layers. Expert classification shows most experts are shared across benign and harmful prompts, though a limited subset shows clear group preference. Top-ranked expert sets show stronger benign-malicious overlap under gradient scores than activation scores, suggesting concentration on a common late-layer expert set. In intervention experiments, suppressing top five benign-dominant experts from activation scores reduces restricted responses from 24 to 14 over 100 prompts, while suppressing gradient-derived experts reduces them from 34 to 22 with fewer unintended reversals. Overall, safety-relevant routing in Mixtral is subtle, depth-dependent, and distributed rather than dominated by a fixed set of experts.
\end{abstract}

\begin{IEEEkeywords}
Mixture of Experts, Mixtral, large language models, routing analysis, safety, harmful prompts, expert selection, activation score, gradient score, interpretability
\end{IEEEkeywords}

\section{Introduction}
Sparse mixture-of-experts (MoE) language models have become increasingly attractive because they increase model capacity without activating the full parameter set for each token. In such models, the router determines which experts are engaged, so routing is not merely an implementation detail but a core part of the model's internal decision process. For safety-sensitive applications, this raises an important question: \emph{does router behavior change in a systematic way when the model is queried with benign versus harmful prompts?}

Mixtral 8x7B is a strong open-weight sparse MoE model in which each transformer layer contains eight experts and the router selects the top two experts for each token \cite{jiang2024mixtral}. The Mixtral paper reports that routing patterns are often more aligned with syntax than domain and that higher layers exhibit stronger temporal locality in consecutive expert assignments \cite{jiang2024mixtral}. Those observations motivate a safety-oriented routing study: even if topic-level specialization is weak, routing may still shift measurably as prompt intent changes from benign to harmful.

In this work, we develop a unified analysis framework for Mixtral internal routing behavior. The framework studies two complementary signal types: activation-based scores, which capture how often experts are selected, and gradient-based scores, which estimate how strongly router parameters influence the sequence loss. For each signal type, we analyze expert behavior and layer behavior at both prompt level and group level. We also introduce expert categorization strategies that compare benign and harmful groups using score gaps and hybrid threshold rules, and we test the resulting expert sets through targeted intervention.

The main contributions of this paper are:
\begin{enumerate}
    \item We present a unified routing-analysis framework for Mixtral using both activation-based and gradient-based scores.
    \item We perform expert-level and layer-level analyses at prompt level and group level for benign and harmful prompt groups.
    \item We show that activation-based routing is broad and long-tailed, whereas gradient-based importance is substantially more concentrated.
    \item We show that activation-based layer selectivity is strongest around layers 8--15, while gradient-based importance becomes sharply concentrated in the final layers.
    \item We classify experts using both activation and gradient signals and show that most experts remain shared across the two prompt groups.
    \item We provide preliminary causal evidence through expert-suppression intervention, showing that both activation-derived and gradient-derived expert sets affect refusal behavior with different trade-offs.
\end{enumerate}

\section{Related Work}
\subsection{Mixture-of-Experts Language Models}
Mixture-of-experts architectures have been studied as a way to scale neural networks through conditional computation \cite{shazeer2017outrageously,lepikhin2020gshard,fedus2022review}. Rather than sending every token through the same feed-forward block, MoE models use a router to select a small subset of experts, thereby increasing total parameter count while keeping per-token computation manageable. Later studies explored routed scaling laws \cite{clark2022unified}, expert-choice routing \cite{zhou2022expertchoice}, and efficient sparse training and inference implementations such as Megablocks \cite{gale2022megablocks}.

Mixtral 8x7B is one of the strongest open-weight sparse MoE models in this line. It uses eight experts per transformer layer and top-2 routing per token, giving each token access to a large sparse parameter pool while activating only a smaller subset of parameters during inference \cite{jiang2024mixtral}. The original paper reports strong performance and includes a routing analysis showing that expert assignment often follows syntactic structure more than semantic domain and that higher layers exhibit stronger expert-assignment repetition across consecutive tokens \cite{jiang2024mixtral}.

\subsection{Routing Analysis and Internal Characterization}
Prior routing analyses of Mixtral mainly examined distributional behavior over datasets and layers rather than safety-oriented prompt categories. The Mixtral paper measures expert-selection proportions across several subsets of The Pile and observes only limited domain specialization, with a mild deviation for DM Mathematics and stronger temporal locality in higher layers \cite{jiang2024mixtral}. These observations motivate our work in two ways. First, they suggest that routing may not separate clearly by broad domain labels, which makes prompt-intent analysis more interesting. Second, they indicate that late layers may be more structured and therefore more promising for safety-oriented analysis.

\subsection{Safety-Relevant Internal Analysis}
Internal safety analysis seeks to understand whether harmful behavior is associated with identifiable internal mechanisms rather than only with output text. In the MoE setting, router behavior offers a natural interface for such analysis because harmful and benign prompts may induce different expert usage patterns, different layer concentration profiles, or different gradient sensitivities. Our work fits into this direction by studying whether harmful prompts shift activation- and gradient-based routing behavior in measurable ways. Unlike purely output-based evaluation, our focus is on characterizing internal distributions and identifying candidate experts and layers for mechanism-oriented study.

\section{Background on Mixtral}
Mixtral 8x7B is a decoder-only sparse MoE model built on a transformer backbone \cite{vaswani2017attention} and the Mistral architecture. It contains 32 transformer layers, each with eight experts replacing the standard feed-forward block, and uses top-2 routing per token \cite{jiang2024mixtral}. The model dimension is 4096, the hidden dimension is 14336, the context length is 32768, and the router selects two experts from the eight available experts at each layer for every token \cite{jiang2024mixtral}. Although the model contains a large sparse parameter count, only a subset is active per token during inference.

Formally, if $x$ is the token representation at a layer, the router computes logits $xW_g$ over experts and keeps only the top-$K$ values, where Mixtral uses $K=2$. The MoE output is the weighted sum of the selected expert outputs \cite{jiang2024mixtral}:
\begin{equation}
    y = \sum_{i=0}^{n-1} \operatorname{Softmax}(\operatorname{Top2}(xW_g))_i \cdot \operatorname{SwiGLU}_i(x).
\end{equation}
This mechanism makes routing directly observable and highly relevant for interpretability: changes in prompt type can potentially change which experts are selected, how concentrated routing is inside each layer, and how influential the corresponding router parameters become. The Mixtral paper further reports that repeated expert assignment across consecutive tokens is substantially higher in layers 15 and 31 than in layer 0, indicating stronger temporal locality in higher layers \cite{jiang2024mixtral}. This motivates our later layer-wise analysis and the hypothesis that late layers may carry stronger structured behavior under harmful prompts.

\section{Methodology}
\subsection{Prompt Groups}
We organize the prompt set into two groups: benign prompts and harmful prompts. In the routing-analysis experiments, each group contains 10 prompts, giving 20 prompts in total. The analysis pipeline supports prompt-level analysis for all 20 prompts individually and group-level analysis by averaging over the benign and harmful sets.

\subsection{Activation-Based Routing Score}
For a given prompt, we attach hooks to the router gate of each MoE layer and record the experts selected by the top-2 routing decision. This produces a raw activation map $A \in \mathbb{R}^{L \times E}$, where $L=32$ is the number of layers and $E=8$ is the number of experts per layer. The raw count $A_{l,e}$ records how many times expert $e$ was selected in layer $l$ across the tokens of that prompt. We then normalize each layer independently:
\begin{equation}
    \tilde{A}_{l,e} = \frac{A_{l,e}}{\sum_{j=1}^{E} A_{l,j}}.
\end{equation}
This normalized map is useful for comparing routing distributions within each layer, although it also implies that each layer contributes fixed total mass 1 to the prompt-wide normalized distribution.

\subsection{Gradient-Based Routing Score}
In addition to activation-based routing, we compute gradient-based expert scores by enabling gradients only on the router gate weights and backpropagating the sequence loss. The gradient score for an expert is based on the average absolute gradient magnitude of the corresponding router gate weights. This provides a complementary notion of expert importance: while activation reveals which experts are selected, gradient reveals which experts are more influential for the model's loss under the given prompt group.

\subsection{Expert-Level Analysis}
For each signal type, we flatten the full $32 \times 8 = 256$ layer--expert pairs for each prompt or group, sort them in descending order by score, and analyze the resulting ranked distribution.

From this ranked list, we study four complementary views. First, we examine the \emph{score-rank distribution}, which shows how quickly the expert scores decay from the largest values to the smallest values. Second, we compute the \emph{cumulative coverage curve}, which measures how much of the total score mass is captured as more ranked layer--expert pairs are included.

To summarize these cumulative curves, we use three coverage-based statistics:
\begin{itemize}
    \item $k80$: the minimum number of top-ranked layer--expert pairs needed to cover 80\% of the total score mass,
    \item $k90$: the minimum number of top-ranked layer--expert pairs needed to cover 90\% of the total score mass,
    \item $k95$: the minimum number of top-ranked layer--expert pairs needed to cover 95\% of the total score mass.
\end{itemize}

These values indicate how concentrated or distributed the expert importance is. Smaller values of $k80$, $k90$, and $k95$ mean that a small number of ranked pairs captures most of the total mass, which implies stronger concentration. Larger values mean that the mass is spread across many layer--expert pairs, which implies a broader distribution.

In addition, we compute an \emph{elbow cutoff}, denoted by $k_{\text{elbow}}$. This value is defined as the rank position at which the largest drop occurs between two consecutive sorted scores. Intuitively, $k_{\text{elbow}}$ provides a data-driven estimate of where the distribution transitions from a small high-score head to a longer low-score tail. While $k80$, $k90$, and $k95$ summarize cumulative concentration, $k_{\text{elbow}}$ highlights the most pronounced structural break in the ranked curve.

\subsection{Layer-Level Analysis}
Because each layer is normalized independently in the activation-based map, prompt-wide normalized layer mass is not directly meaningful as a layer-importance measure. Instead, we study each layer using concentration and spread metrics:
\begin{itemize}
    \item \textbf{dominant score}: highest expert score in a layer,
    \item \textbf{top-2 sum}: sum of the two largest expert scores,
    \item \textbf{entropy}:
    \begin{equation}
        H = -\sum_{i=1}^{E} p_i \log p_i,
    \end{equation}
    \item \textbf{effective experts}:
    \begin{equation}
        E_{\text{eff}} = e^H,
    \end{equation}
    \item \textbf{active expert count}: number of experts with nonzero usage.
\end{itemize}
The same style of layer-wise concentration analysis is applied to the gradient-based score distributions.


\subsection{Expert Classification}
To move from descriptive score distributions to interpretable expert roles, we classify each layer--expert pair using the difference between the malicious-group average score and the benign-group average score, together with a minimum average-magnitude constraint.

For the activation-based classification, each layer--expert pair is associated with two group-level averages: \texttt{benign\_avg\_score} and \texttt{malicious\_avg\_score}. For the gradient-based classification, the corresponding quantities are \texttt{benign\_avg\_grad} and \texttt{malicious\_avg\_grad}. In both cases, we define the \emph{safety gap} as the difference between the malicious-group average and the benign-group average:
\begin{equation}
\text{safety\_gap} = \text{malicious average} - \text{benign average}.
\end{equation}
A positive safety gap means that the expert is more active or more influential for malicious prompts than for benign prompts, while a negative safety gap means the opposite. We also compute the absolute value of this difference,
\begin{equation}
\text{abs\_gap} = |\text{safety\_gap}|,
\end{equation}
which measures the strength of the separation regardless of direction.

The classification rule also uses two user-chosen thresholds: a \emph{gap threshold} and a \emph{minimum average magnitude}. The gap threshold controls how large the benign--malicious difference must be before an expert is treated as group-preferential. The minimum average magnitude ensures that an expert is not labeled as important only because of a numerical difference between two very small values.

For activation-based classification, we set
\begin{equation}
\text{gap\_threshold} = 0.05, \qquad
\text{min\_avg\_magnitude} = 0.08.
\end{equation}
These values were selected heuristically by inspecting the \texttt{safety\_gap}, \texttt{benign\_avg\_score}, and \texttt{malicious\_avg\_score} columns in the activation classification file and choosing thresholds that isolate a small set of experts with visibly large positive or negative separation. In other words, the threshold 0.05 was chosen to focus on experts with relatively large activation-score differences, while 0.08 was chosen by examining the two average-score columns and excluding experts whose average scores were too small to be practically meaningful.

For gradient-based classification, we set
\begin{equation}
\text{gap\_threshold} = 10^{-4}, \qquad
\text{min\_avg\_magnitude} = 10^{-4}.
\end{equation}
These values were again selected heuristically by inspecting the \texttt{abs\_gap}, \texttt{benign\_avg\_grad}, and \texttt{malicious\_avg\_grad} columns in the gradient classification file and choosing thresholds that highlight only a small number of clearly separated experts. Because gradient values are much smaller in magnitude than activation scores, the gradient thresholds are correspondingly smaller.

Using these quantities, each layer--expert pair is assigned to one of six categories:
\begin{itemize}
    \item \textbf{malicious-dominant}: the safety gap is at least the threshold and the malicious-group average is large enough,
    \item \textbf{benign-dominant}: the safety gap is at most the negative threshold and the benign-group average is large enough,
    \item \textbf{shared}: the absolute gap is smaller than the threshold,
    \item \textbf{weak-malicious}: the gap is positive and large enough, but the malicious-group average is still below the minimum magnitude,
    \item \textbf{weak-benign}: the gap is negative and large enough, but the benign-group average is still below the minimum magnitude,
    \item \textbf{uncertain}: any remaining case that does not match the previous rules.
\end{itemize}


This classification should therefore be interpreted as a heuristic filtering rule rather than a statistically optimized classifier. Its purpose is to separate strongly group-preferential experts from weak or nearly shared experts in a transparent and reproducible way.

\subsection{Expert Intervention}
To test whether the experts identified by the routing analysis are mechanistically relevant, we suppress the top five benign-dominant layer--expert pairs identified by each classification method during generation. Each intervention experiment compares baseline and suppressed responses on 100 harmful prompts. We then label each paired output semantically as \emph{restricted} if the model refuses or otherwise does not meaningfully comply with the harmful intent, and \emph{non-restricted} if it provides harmful guidance or substantial assistance. This produces a simple paired evaluation of whether expert suppression weakens refusal behavior.

\subsection{Presentation Strategy}
For brevity, the main paper presents only group-level figures, while prompt-level analyses are summarized through numerical tables and described in the text. This reduces figure redundancy while preserving the prompt-level evidence.

\section{Experimental Setup}
The experimental code supports 19 lab modes and unifies the full analysis workflow \cite{analysiscode}. For activation-based analysis, Labs 1--12 cover prompt-level expert ranking, expert classification, full-distribution analysis, group-level aggregation, and layer-wise analysis. For gradient-based analysis, Labs 13--19 replicate the same structure using gradient scores instead of activation scores \cite{analysiscode}.

The present paper uses the completed outputs from both parts of the pipeline. For activation-based analysis, we use prompt-level and group-level expert summaries, group-level expert-distribution figures, layer-wise summary tables, and group-level layer-wise figures. For gradient-based analysis, we use prompt-level and group-level expert summaries, group-level gradient-distribution figures, layer-wise gradient summary tables, group-level gradient layer-wise figures, and expert-classification results. Finally, we use two intervention result files, one produced by suppressing the activation-derived expert set and one produced by suppressing the gradient-derived expert set.

\section{Results}
For brevity, we present only the group-level figures in the main paper. The prompt-level analyses are still used in the quantitative summaries and are discussed in the text, but the full prompt-level figure grids can be moved to supplementary material. This keeps the main narrative focused on the dominant trends while preserving the prompt-level evidence through summary tables.

\subsection{Expert-Level Analysis}
We first analyze expert-level routing behavior using two complementary signals: activation-based scores and gradient-based scores. The activation score reflects how often an expert is selected by the router, while the gradient score reflects how strongly the router parameters associated with that expert influence the sequence loss. These two signals provide different but related views of expert importance.

\subsubsection{Activation-Based Expert Analysis}
At prompt level, the activation-based routing distributions are broad and long-tailed. The prompt-level significant-expert summary in Table~\ref{tab:prompt_activation_summary} shows that, on average, 151.1 benign and 154.3 harmful ranked layer--expert pairs are required to cover 80\% of the cumulative normalized routing mass, while the corresponding averages for 90\% and 95\% coverage are 185.6/188.7 and 209.6/212.9. Since the full space contains 256 layer--expert pairs, these values indicate that activation-based routing is distributed across a large portion of the model rather than collapsing to only a few experts.

\begin{figure*}[t]
    \centering
    \begin{subfigure}[b]{0.47\textwidth}
        \centering
        \includegraphics[width=\textwidth]{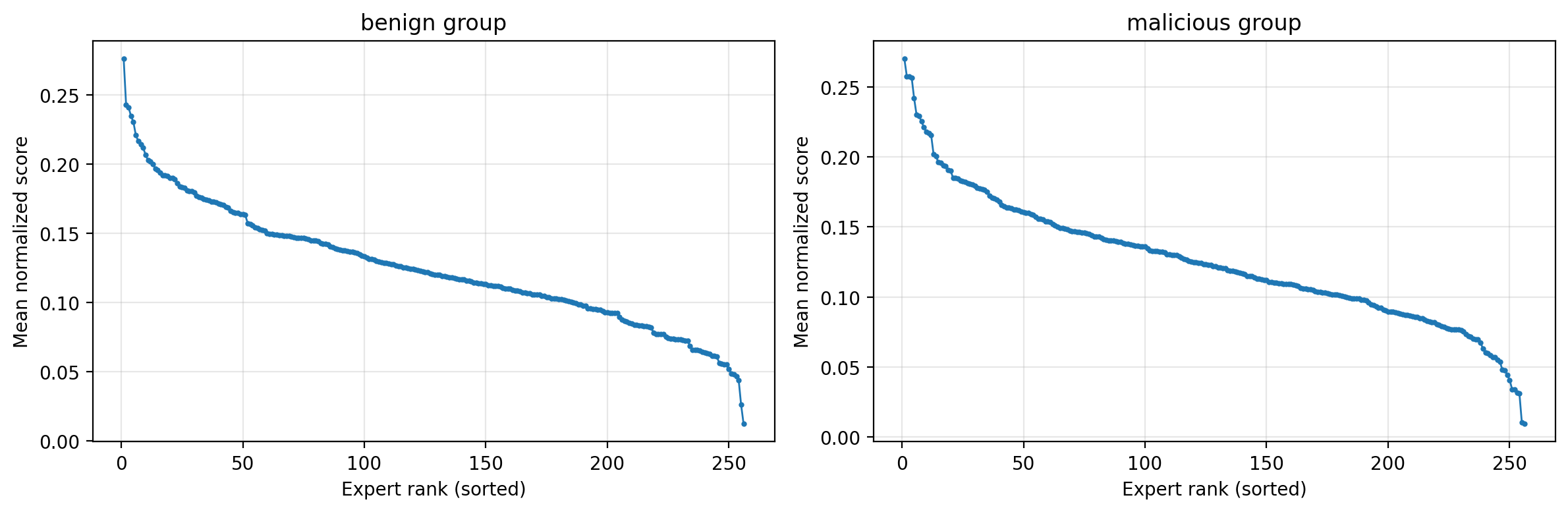}
        \caption{}
    \end{subfigure}
    \hfill
    \begin{subfigure}[b]{0.47\textwidth}
        \centering
        \includegraphics[width=\textwidth]{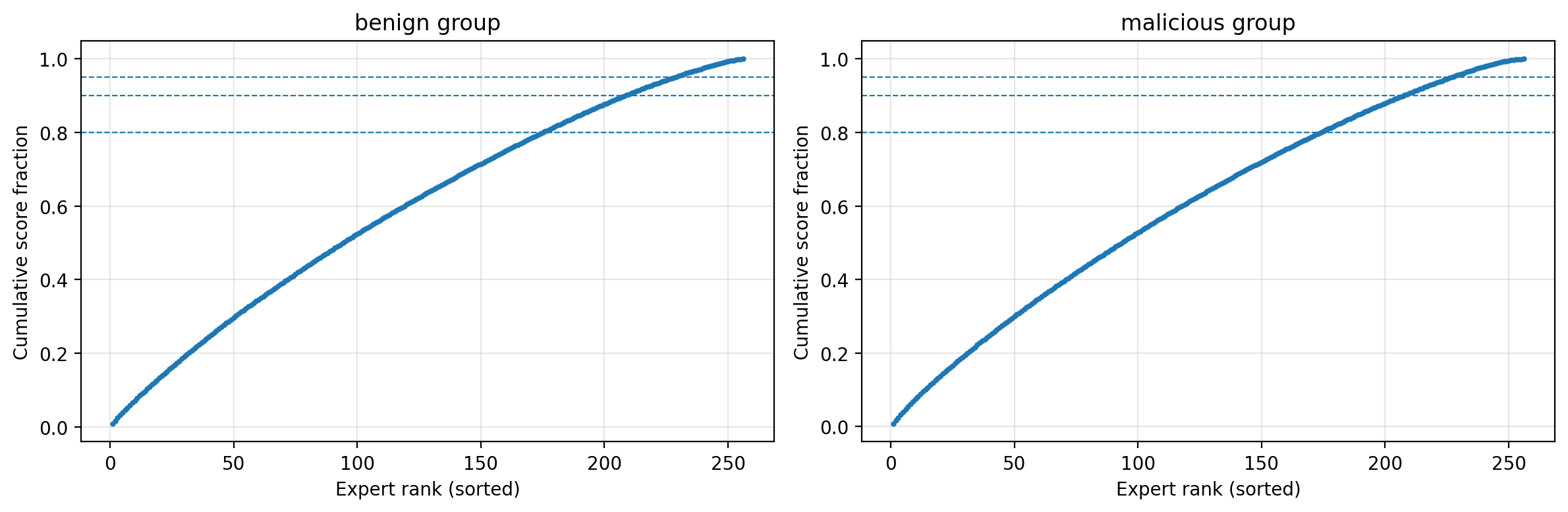}
        \caption{}
    \end{subfigure}
    \caption{Group-level activation-based expert analysis. Left: sorted mean activation score distribution. Right: cumulative activation score distribution. The harmful group is slightly more concentrated, but the two groups remain broadly similar overall.}
    \label{fig:group_activation_expert}
\end{figure*}

At group level, the benign and harmful activation distributions remain close overall, as shown in Fig.~\ref{fig:group_activation_expert}. The harmful group is only slightly more concentrated: its top-5 cumulative score is 1.283 versus 1.226 for the benign group, and its group-level coverage values are slightly smaller ($k80=174$ versus 176, $k90=208$ versus 209, and $k95=227$ versus 229). Thus, activation-based expert routing shows only modest separation between benign and harmful prompts.

\begin{table}[t]
    \centering
    \caption{Prompt-level significant-expert summary from activation scores. Values shown are group means over 10 prompts.}
    \label{tab:prompt_activation_summary}
    \footnotesize
    \setlength{\tabcolsep}{4pt}
    \begin{tabular}{lcccccc}
        \toprule
        Group & $k80$ & $k90$ & $k95$ & $k_{\text{elbow}}$ & top1 & top5 \\
        \midrule
        Benign & 151.1 & 185.6 & 209.6 & 6.3 & 0.367 & 1.647 \\
        Harmful & 154.3 & 188.7 & 212.9 & 8.3 & 0.340 & 1.543 \\
        \bottomrule
    \end{tabular}
\end{table}

\begin{table}[t]
    \centering
    \caption{Group-level significant-expert summary from activation scores.}
    \label{tab:group_activation_summary}
    \footnotesize
    \setlength{\tabcolsep}{4pt}
    \begin{tabular}{lcccccc}
        \toprule
        Group & $k80$ & $k90$ & $k95$ & $k_{\text{elbow}}$ & top1 & top5 \\
        \midrule
        Benign & 176 & 209 & 229 & 1 & 0.276 & 1.226 \\
        Harmful & 174 & 208 & 227 & 254 & 0.270 & 1.283 \\
        \bottomrule
    \end{tabular}
\end{table}


\subsubsection{Gradient-Based Expert Analysis}
The gradient-based expert results show a much sharper distribution. At prompt level, the significant-expert summary in Table~\ref{tab:prompt_gradient_summary} indicates that only 52.0 benign and 57.9 harmful ranked pairs are needed to cover 80\% of the cumulative gradient mass on average, compared with 151.1 and 154.3 in the activation-based case. Likewise, the mean 95\% coverage values are 122.5 and 128.0, far below the activation-based values. This demonstrates that gradient-based importance is much more concentrated than activation-based routing frequency.

\begin{figure*}[t]
    \centering
    \begin{subfigure}[b]{0.47\textwidth}
        \centering
        \includegraphics[width=\textwidth]{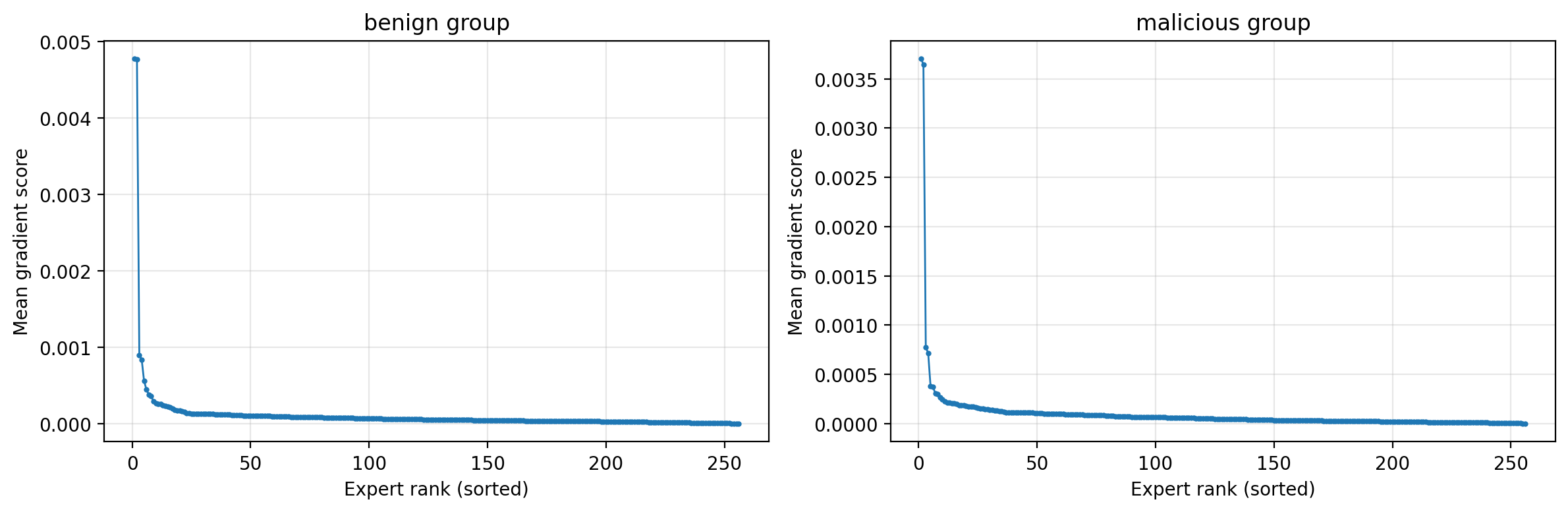}
        \caption{}
    \end{subfigure}
    \hfill
    \begin{subfigure}[b]{0.47\textwidth}
        \centering
        \includegraphics[width=\textwidth]{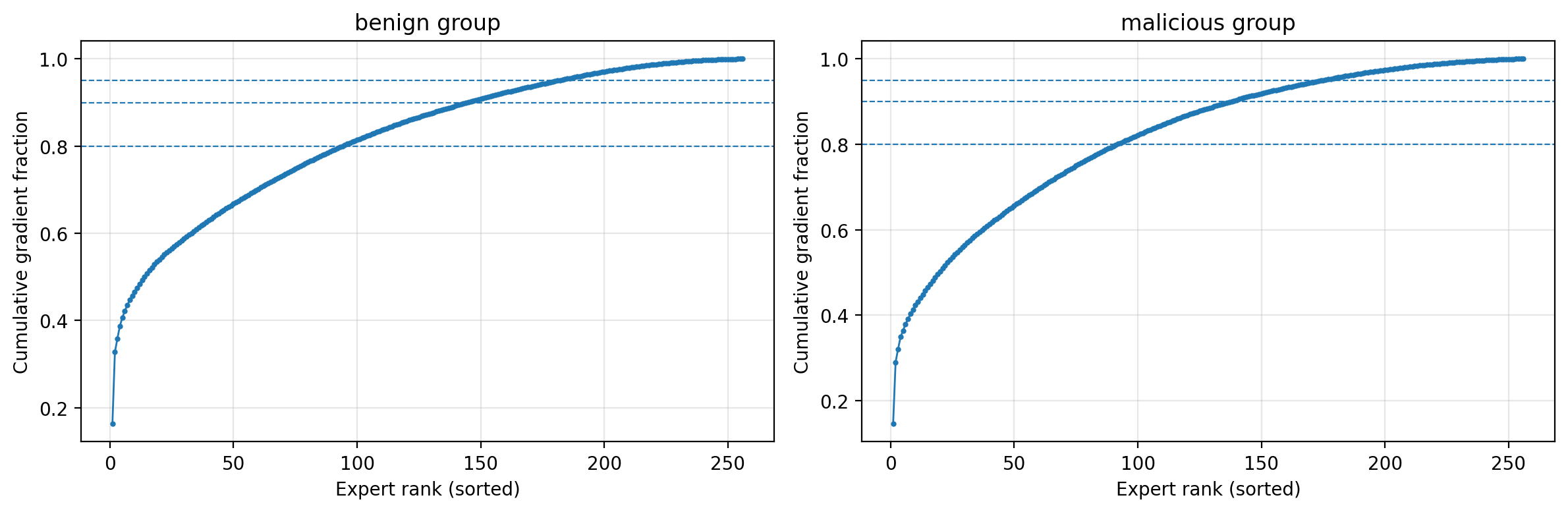}
        \caption{}
    \end{subfigure}
    \caption{Group-level gradient-based expert analysis. Left: sorted mean gradient score distribution. Right: cumulative gradient score distribution. Both groups are much more concentrated than in the activation-based case.}
    \label{fig:group_gradient_expert}
\end{figure*}

At group level, the gradient distributions remain similar across benign and harmful prompts, but both are substantially steeper than the activation-based curves, as shown in Fig.~\ref{fig:group_gradient_expert}. The harmful group becomes slightly more concentrated after averaging, with $k80=92$ versus 94, $k90=138$ versus 145, and $k95=175$ versus 182. However, the head values remain slightly larger for the benign group, with top-1 gradient 0.004772 versus 0.003703 and top-5 gradient sum 0.011839 versus 0.009222. This indicates that harmful prompts do not simply induce a single stronger peak; rather, the difference lies in the overall shape of the ranked gradient distribution.

\begin{table}[t]
    \centering
    \caption{Prompt-level significant-expert summary from gradient scores. Values shown are group means over 10 prompts.}
    \label{tab:prompt_gradient_summary}
    \footnotesize
    \setlength{\tabcolsep}{4pt}
    \begin{tabular}{lcccccc}
        \toprule
        Group & $k80$ & $k90$ & $k95$ & $k_{\text{elbow}}$ & top1 & top5 \\
        \midrule
        Benign & 52.0 & 89.0 & 122.5 & 2.0 & 0.0048 & 0.0122 \\
        Harmful & 57.9 & 94.8 & 128.0 & 2.0 & 0.0037 & 0.0096 \\
        \bottomrule
    \end{tabular}
\end{table}

\begin{table}[t]
    \centering
    \caption{Group-level significant-expert summary from gradient scores.}
    \label{tab:group_gradient_summary}
    \footnotesize
    \setlength{\tabcolsep}{4pt}
    \begin{tabular}{lcccccc}
        \toprule
        Group & $k80$ & $k90$ & $k95$ & $k_{\text{elbow}}$ & top1 & top5 \\
        \midrule
        Benign & 94 & 145 & 182 & 2 & 0.004772 & 0.011839 \\
        Harmful & 92 & 138 & 175 & 2 & 0.003703 & 0.009222 \\
        \bottomrule
    \end{tabular}
\end{table}


\subsubsection{Expert-Level Interpretation}
Taken together, the expert-level analyses show that activation and gradient provide different notions of importance. Activation-based routing is broad and distributed across many layer--expert pairs, whereas gradient-based importance is much more concentrated. However, in both views the benign and harmful groups remain close overall. Therefore, the current evidence does not support a claim that a tiny fixed set of experts uniquely explains harmful behavior. Instead, expert-level differences appear subtle, distributed, and dependent on the analysis signal.

\subsection{Layer-Level Analysis}
We next analyze layer behavior using layer-wise concentration and spread metrics. For activation-based analysis, we use dominant score, top-2 sum, entropy, effective experts, and active expert count computed from the normalized routing distribution within each layer. For gradient-based analysis, we compute the analogous metrics from the layer-wise gradient score distribution.

\subsubsection{Activation-Based Layer Analysis}
\begin{figure*}[t]
    \centering
    \begin{subfigure}[b]{0.47\textwidth}
        \centering
        \includegraphics[width=\textwidth]{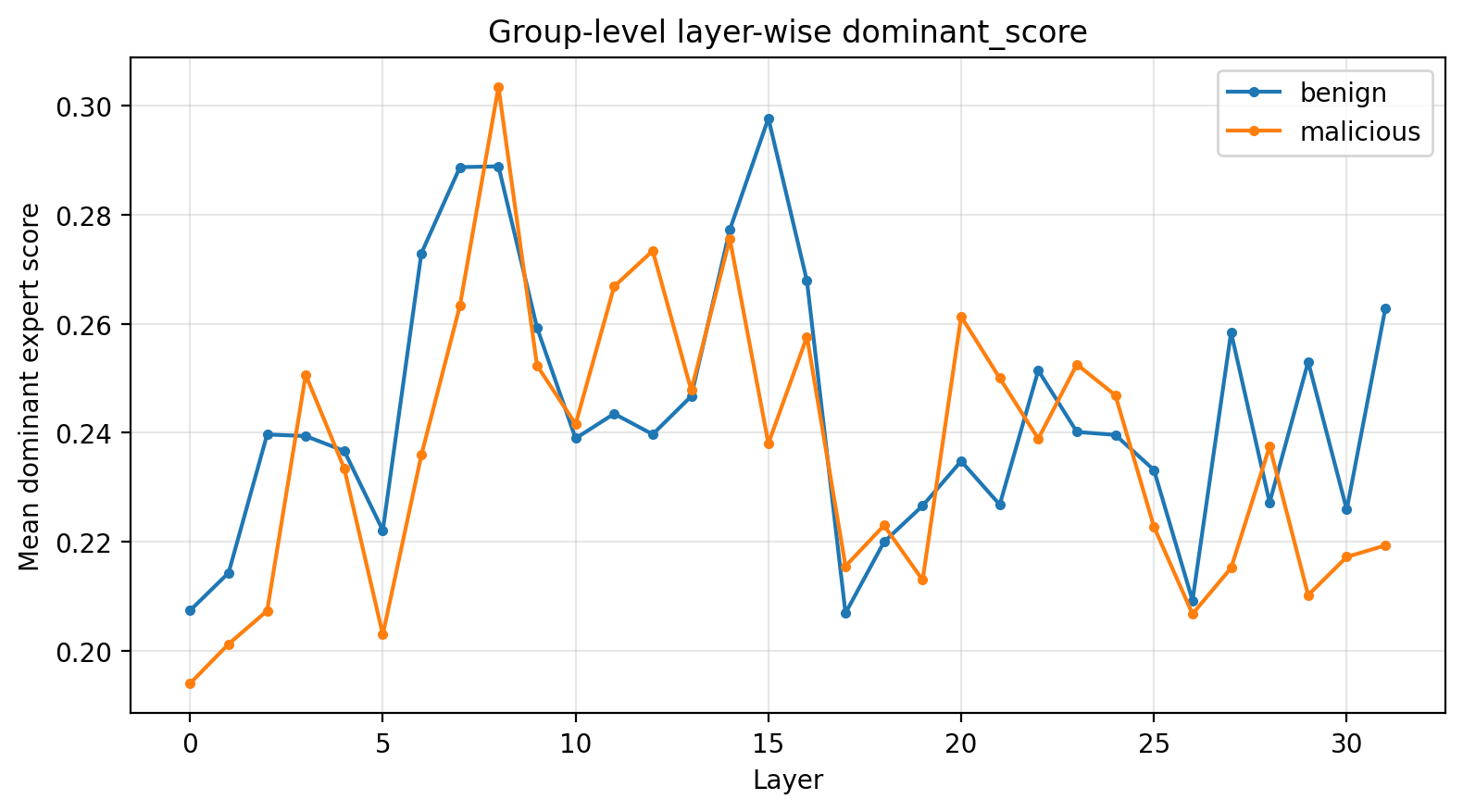}
        \caption{}
    \end{subfigure}
    \hfill
    \begin{subfigure}[b]{0.47\textwidth}
        \centering
        \includegraphics[width=\textwidth]{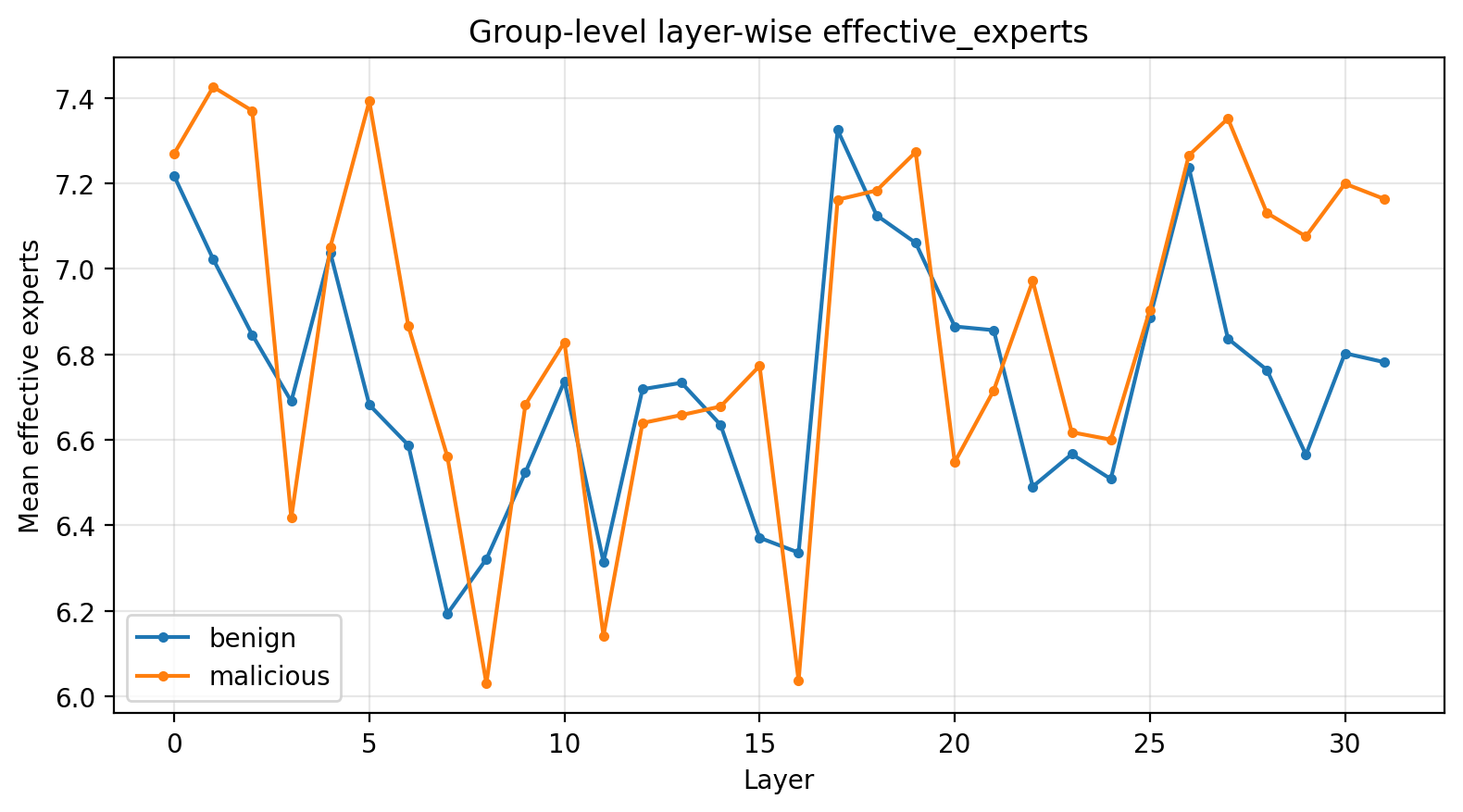}
        \caption{}
    \end{subfigure}
    \caption{Group-level activation-based layer analysis. Left: mean dominant expert score by layer. Right: mean effective experts by layer. The strongest concentration occurs around layers 8--15, while the harmful group becomes more diffuse in the final layers.}
    \label{fig:group_activation_layer}
\end{figure*}

The prompt-level activation layer curves show that dominant score varies across prompts and layers, but there is no consistent monotonic increase toward the end of the network. At group level, both benign and harmful prompts are most concentrated in the middle-to-late region around layers 8--15, as shown in Fig.~\ref{fig:group_activation_layer}. The benign group reaches its maximum dominant score of 0.298 at layer 15, while the harmful group reaches its maximum dominant score of 0.303 at layer 8. In the same region, the effective-expert count becomes relatively small, with minimum values 6.193 for the benign group and 6.030 for the harmful group.

The overall group statistics further show that harmful prompts are not more concentrated under the activation-based view. Averaged over all 32 layers, the benign group has higher mean dominant score (0.244 versus 0.237), higher mean top-2 sum (0.433 versus 0.424), and lower mean effective-expert count (6.739 versus 6.874). This difference becomes more visible in the final block of layers 24--31, where the harmful group becomes more diffuse rather than more concentrated.

\begin{table}[t]
    \centering
    \caption{Group-level layer-wise activation summary over all 32 layers.}
    \label{tab:layer_activation_summary}
    \footnotesize
    \setlength{\tabcolsep}{4pt}
    \begin{tabular}{lcccc}
        \toprule
        Group & Dom. & Eff. & Top-2 & Ent. \\
        \midrule
        Benign & 0.244 & 6.739 & 0.433 & 0.915 \\
        Harmful & 0.237 & 6.874 & 0.424 & 0.925 \\
        \bottomrule
    \end{tabular}
\end{table}

\begin{table}[t]
    \centering
    \caption{Late-layer activation summary over layers 24--31.}
    \label{tab:late_layer_activation_summary}
    \footnotesize
    \setlength{\tabcolsep}{4pt}
    \begin{tabular}{lcccc}
        \toprule
        Group & Dom. & Eff. & Top-2 & Ent. \\
        \midrule
        Benign & 0.239 & 6.798 & 0.423 & 0.920 \\
        Harmful & 0.222 & 7.086 & 0.403 & 0.940 \\
        \bottomrule
    \end{tabular}
\end{table}


\subsubsection{Gradient-Based Layer Analysis}
\begin{figure*}[t]
    \centering
    \begin{subfigure}[b]{0.47\textwidth}
        \centering
        \includegraphics[width=\textwidth]{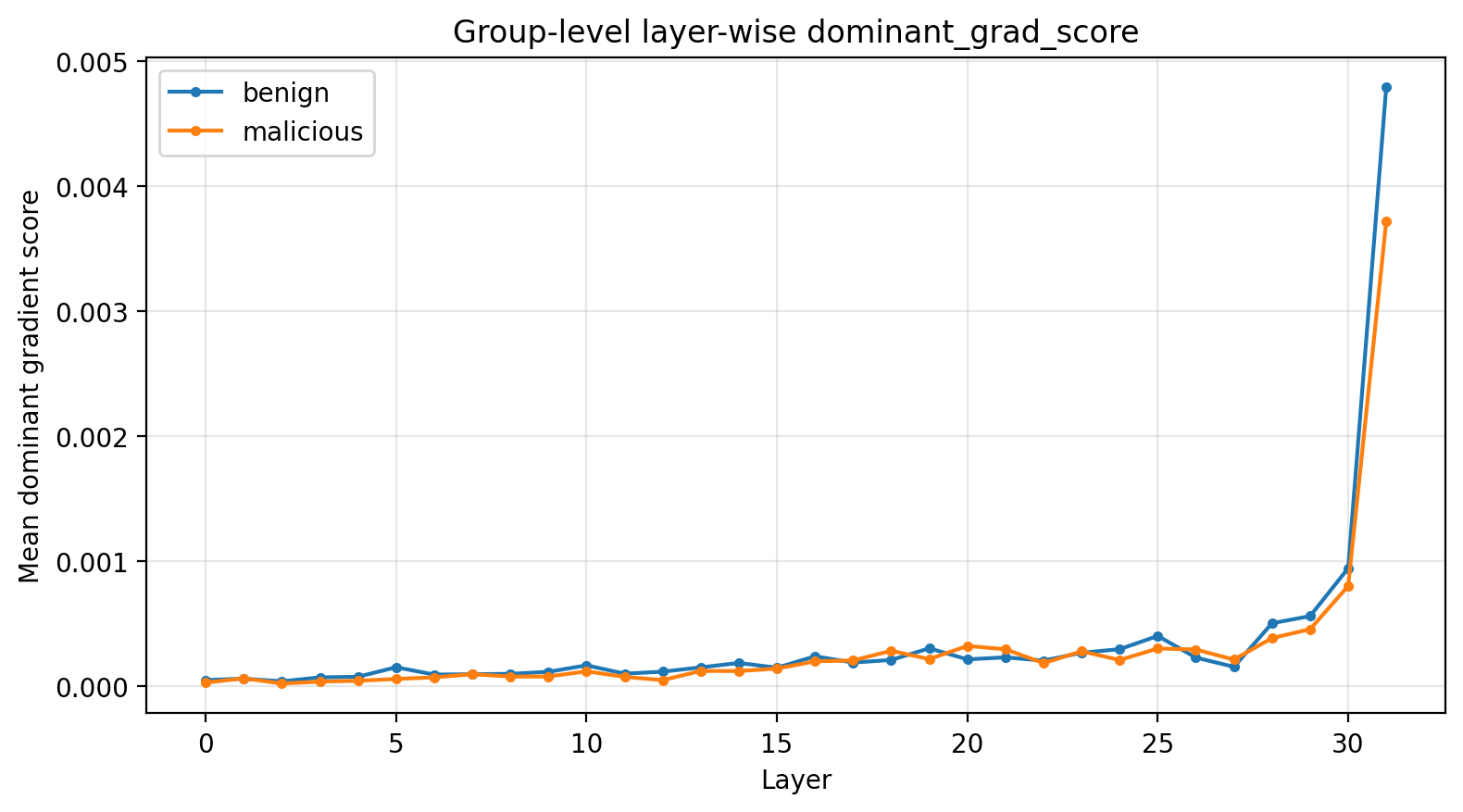}
        \caption{}
    \end{subfigure}
    \hfill
    \begin{subfigure}[b]{0.47\textwidth}
        \centering
        \includegraphics[width=\textwidth]{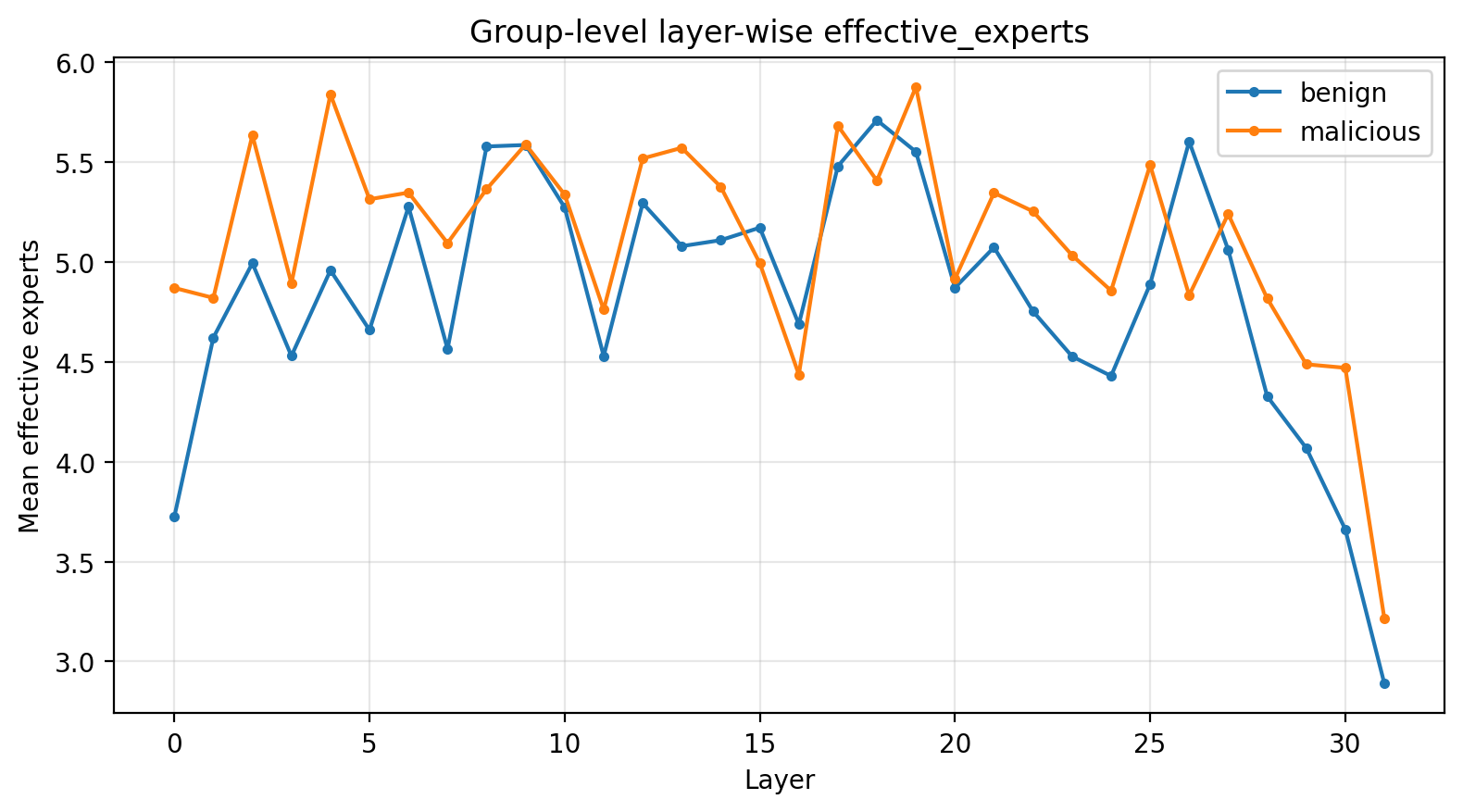}
        \caption{}
    \end{subfigure}
    \caption{Group-level gradient-based layer analysis. Left: mean dominant gradient score by layer. Right: mean effective experts by layer. Both groups become sharply more concentrated in the final layers, with the strongest peak at layer 31.}
    \label{fig:group_gradient_layer}
\end{figure*}

The gradient-based layer analysis reveals a much sharper depth effect. At prompt level, dominant gradient scores are generally small in early layers and become larger in later layers, while the effective-expert counts are noticeably smaller than in the activation-based case. At group level, the dominant gradient score remains relatively small through early and middle layers, then rises sharply toward the end of the network, with the strongest peak at the final layer, as shown in Fig.~\ref{fig:group_gradient_layer}. For the benign group, the mean dominant gradient score increases from $5.15\times10^{-5}$ at layer 0 to $4.79\times10^{-3}$ at layer 31. For the harmful group, it increases from $3.11\times10^{-5}$ to $3.72\times10^{-3}$.

The effective-expert curves show the same pattern from the concentration perspective. Through most of the network, both groups remain in the range of roughly 4.5--5.9 effective experts, already lower than the activation-based values. In the final layers, the effective-expert count drops further to 2.89 for the benign group and 3.22 for the harmful group. Thus, gradient-based importance becomes highly concentrated in the final block.

\begin{table}[t]
    \centering
    \caption{Group-level layer-wise gradient summary over all 32 layers.}
    \label{tab:layer_gradient_summary}
    \footnotesize
    \setlength{\tabcolsep}{4pt}
    \begin{tabular}{lcccc}
        \toprule
        Group & Dom. & Eff. & Top-2 & Ent. \\
        \midrule
        Benign & $3.60{\times}10^{-4}$ & 4.829 & $6.74{\times}10^{-4}$ & 0.745 \\
        Harmful & $3.01{\times}10^{-4}$ & 5.115 & $5.58{\times}10^{-4}$ & 0.774 \\
        \bottomrule
    \end{tabular}
\end{table}

\begin{table}[t]
    \centering
    \caption{Late-layer gradient summary over layers 24--31.}
    \label{tab:late_layer_gradient_summary}
    \footnotesize
    \setlength{\tabcolsep}{4pt}
    \begin{tabular}{lcccc}
        \toprule
        Group & Dom. & Eff. & Top-2 & Ent. \\
        \midrule
        Benign & $9.87{\times}10^{-4}$ & 4.366 & $1.897{\times}10^{-3}$ & 0.693 \\
        Harmful & $7.99{\times}10^{-4}$ & 4.675 & $1.516{\times}10^{-3}$ & 0.727 \\
        \bottomrule
    \end{tabular}
\end{table}


\subsubsection{Layer-Level Interpretation}
The layer-level comparison highlights the key difference between activation and gradient. Under activation, the most selective region is the middle-to-late portion around layers 8--15, and harmful prompts become slightly more diffuse in the final layers. Under gradient, however, the strongest concentration appears at the end of the network, and the final layers carry disproportionately large loss-sensitive importance. Therefore, the activation signal and gradient signal point to different kinds of structure: activation captures broad routing preference, whereas gradient isolates the layers where routing decisions matter most for the loss.

\subsection{Expert Classification}
To move from descriptive score distributions to interpretable expert roles, we classify each layer--expert pair using the benign--harmful score gap together with a minimum average-magnitude constraint. Under the activation-based classification, 216 of the 256 layer--expert pairs are shared, 22 are benign-dominant, and 18 are malicious-dominant. Under the gradient-based classification, the distribution is much sparser: 245 are shared, 6 are benign-dominant, and 5 are malicious-dominant. Therefore, activation identifies a broader set of group-preferential experts, whereas gradient isolates a smaller but sharper set of candidate influential experts.

The strongest malicious-dominant activation experts include layer 16 expert 7, layer 12 expert 2, layer 6 expert 3, and layer 3 expert 6. The strongest benign-dominant activation experts include layer 7 expert 5, layer 12 expert 1, layer 15 expert 1, and layer 27 expert 3. For the gradient-based classification, the most prominent malicious-dominant experts include layer 21 expert 6, layer 18 expert 3, layer 26 expert 0, layer 20 expert 6, and layer 20 expert 1, while the strongest benign-dominant gradient experts include layer 29 expert 4, layer 28 expert 5, layer 26 expert 7, layer 19 expert 6, and layer 24 expert 1.

An additional comparison between the classification outputs and the raw group-level ranking files reveals an important difference between activation and gradient. In the classification CSVs, the top benign-dominant and top malicious-dominant layer--expert pairs do not overlap, which is expected because the classification procedure separates experts according to the sign of the benign--harmful score gap. However, when we examine the raw group-level top-10 ranked layer--expert pairs before classification, the gradient-based scores show strong overlap between the benign and malicious groups, whereas the activation-based scores show only limited overlap. In the gradient ranking file, eight of the top-10 layer--expert pairs are shared by both groups: (31,2), (31,7), (30,6), (30,4), (29,4), (29,3), (28,2), and (28,4). In contrast, the activation ranking file shows only three shared top-10 pairs: (14,2), (13,0), and (8,3).

This difference is informative. It suggests that the gradient-based signal concentrates both benign and malicious prompts onto a very similar high-importance set in the late layers, even though the final gradient-based classification still separates a small subset of experts as benign-dominant or malicious-dominant based on score differences. By contrast, the activation-based signal is more distributed and yields less overlap in the raw top-ranked experts, which is consistent with the earlier observation that activation-based routing remains broad and long-tailed.

These two views are therefore complementary. Activation-based classification reflects repeated routing preference across many tokens and prompts, whereas gradient-based classification reflects where the loss is most sensitive to router behavior. The ranking comparison further shows that gradient-based importance is not only more concentrated, but also more shared at the top across benign and malicious groups. In both cases, however, the majority of experts remain shared overall, which reinforces the broader conclusion that safety-relevant routing behavior in Mixtral is subtle and distributed rather than reducible to a single fixed harmful-expert set.

\subsection{Expert Intervention Based on Activation- and Gradient-Derived Expert Sets}
To test whether the experts identified by the routing analysis are mechanistically relevant, we performed two targeted intervention experiments. In the first experiment, we suppressed the top five benign-dominant layer--expert pairs identified from the activation-based expert classification. In the second experiment, we suppressed the top five benign-dominant layer--expert pairs identified from the gradient-based expert classification.

For each harmful prompt, we generated two outputs: a baseline response and a suppressed response. We then compared the paired outputs using a binary semantic label: \emph{restricted} if the model refused or otherwise did not meaningfully comply with the harmful intent, and \emph{non-restricted} if it provided harmful guidance or substantial assistance.

For the activation-based intervention, 100 paired outputs were evaluated. Under baseline generation, 24 responses were classified as restricted, and after suppressing the selected activation-based experts this number decreased to 14. Thus, the intervention reduced the restricted-response count by 10 prompts. At transition level, 18 prompts changed from restricted to non-restricted after suppression, while 8 prompts changed in the opposite direction. The remaining 73 prompts preserved the same category, including 68 that were non-restricted in both conditions and 6 that remained restricted in both conditions.

For the gradient-based intervention, 100 paired outputs were evaluated. Under baseline generation, 34 responses were classified as restricted, and after suppressing the selected gradient-based experts this number decreased to 22. Thus, the intervention reduced the restricted-response count by 12 prompts. At transition level, 15 prompts changed from restricted to non-restricted after suppression, while only 3 prompts changed from non-restricted to restricted. The remaining 82 prompts preserved the same category, including 63 that were non-restricted in both conditions and 19 that remained restricted in both conditions.

\begin{table}[t]
    \centering
    \caption{Comparison of expert-suppression interventions.}
    \label{tab:intervention_summary}
    \footnotesize
    \setlength{\tabcolsep}{4pt}
    \begin{tabular}{lcc}
        \toprule
        Metric & Activation & Gradient \\
        \midrule
        Number of prompts & 100 & 100 \\
        Baseline restricted & 24 & 34 \\
        Suppressed restricted & 14 & 22 \\
        Restricted $\rightarrow$ non-restricted & 18 & 15 \\
        Non-restricted $\rightarrow$ restricted & 8 & 3 \\
        Restricted in both & 6 & 19 \\
        Non-restricted in both & 68 & 63 \\
        \bottomrule
    \end{tabular}
\end{table}

These results suggest that both intervention strategies identify experts related to refusal behavior, but they exhibit different trade-offs. The activation-based intervention yields a larger relative reduction in restricted responses, decreasing the refusal count from 24 to 14, which corresponds to about a 41.7\% reduction relative to baseline. The gradient-based intervention yields a slightly larger absolute reduction in restricted responses, decreasing the refusal count from 34 to 22, but its relative reduction is smaller at about 35.3\%. At the same time, the gradient-based intervention appears more stable, because it produces fewer reverse-direction changes: only 3 prompts move from non-restricted to restricted, compared with 8 for the activation-based intervention.

Therefore, neither intervention can yet be considered uniformly superior. The activation-based expert set appears somewhat stronger in terms of relative refusal reduction, whereas the gradient-based expert set appears cleaner in terms of fewer unintended reversals. Taken together, these results provide preliminary causal evidence that the experts identified by both classification methods contribute to observable safety behavior, while also indicating that refusal behavior is not controlled by only one small expert subset.

A direct comparison between the two intervention settings should still be interpreted cautiously, since the paired outputs come from separate runs and the baseline restricted counts differ. Nevertheless, both experiments indicate that suppressing the selected benign-dominant experts weakens refusal behavior for a subset of prompts, providing preliminary mechanistic support for the expert-classification analysis.

\section{Discussion}
The current findings align partially with the original Mixtral routing study. The Mixtral paper reported that routing patterns are often more aligned with syntax than with broad semantic domain and that higher layers exhibit stronger temporal locality. Our results are consistent with the view that global routing differences are subtle. Even when prompt intent changes from benign to harmful, the overall routing distributions remain broad or overlapping at group level, depending on the analysis signal.

At the same time, our study extends prior routing analysis in a safety-oriented direction. Rather than comparing large corpus domains, we compare benign and harmful prompt categories and examine whether safety-relevant behavior is reflected in expert usage, layer concentration, and intervention response. The resulting differences are measurable, but they do not indicate that harmful behavior is controlled by one universal harmful-expert subset. Instead, the evidence points to a more distributed mechanism involving shifts across both experts and layers.

The contrast between activation and gradient signals is especially informative. Activation-based analysis captures repeated routing preference and shows a broad, long-tailed distribution across layer--expert pairs. Gradient-based analysis captures loss sensitivity and yields a much sharper distribution. This difference also appears in the raw top-ranked expert sets: the gradient top-10 lists for benign and harmful groups overlap strongly, whereas the activation top-10 lists overlap only weakly. This suggests that gradient importance concentrates both prompt groups onto a more similar late-layer core, while activation remains more distributed across experts.

The layer-level results reinforce this distinction. Under activation, the most selective region appears in the middle-to-late part of the network, roughly layers 8--15, and harmful prompts become slightly more diffuse in the final layers. Under gradient, however, the strongest concentration appears at the end of the network, where the dominant gradient score rises sharply and the effective number of experts drops noticeably. Thus, activation and gradient expose different aspects of internal model behavior: activation reflects broad routing preference, whereas gradient highlights the layers where routing decisions matter most for the loss.

The intervention experiments provide preliminary causal support for this interpretation. Suppressing experts identified from either activation-based or gradient-based classification reduces the number of restricted responses, indicating that the selected experts are related to observable refusal behavior. However, neither intervention produces a universal behavioral collapse. The activation-derived expert set yields a stronger relative reduction in restricted responses, whereas the gradient-derived expert set produces fewer unintended reversals. This difference suggests that the two expert-selection strategies capture related but non-identical aspects of the internal safety mechanism.

Overall, the results suggest that safety-relevant routing in Mixtral is distributed, depth-dependent, and signal-dependent. For MoE safety analysis, relying on only one internal metric would provide an incomplete picture. A more informative approach is to jointly analyze routing frequency, loss-sensitive importance, and intervention behavior.

\section{Conclusion}
This paper presented a safety-oriented routing analysis of Mixtral 8x7B-Instruct under benign and harmful prompts. We developed a unified framework that studies expert-level and layer-level internal behavior using activation and gradient signals, and we connected these descriptive analyses to targeted intervention experiments. The results show that activation-based routing is broad and distributed, gradient-based importance is much more concentrated, activation-based layer selectivity is strongest around layers 8--15, and gradient-based importance becomes sharply focused in the final layers.

Expert classification further shows that most experts are shared across benign and harmful prompts, although a limited subset exhibits clear group preference under each signal. Beyond the classification labels themselves, the raw ranking comparison reveals that gradient-based top experts are much more strongly shared across benign and harmful groups than activation-based top experts, indicating that gradient importance is concentrated on a more common late-layer expert set.

The intervention experiments provide preliminary causal evidence that experts selected by both classification methods contribute to observable refusal behavior. At the same time, neither intervention produces a complete behavioral shift, which supports the broader conclusion that safety-relevant routing in Mixtral is not controlled by a single small fixed expert subset. Instead, the mechanism appears to be subtle, distributed, and dependent on both depth and analysis signal.

More generally, the contrast between activation and gradient demonstrates that different internal metrics reveal different aspects of MoE safety behavior. Activation captures broad routing preference, gradient highlights loss-sensitive concentration, and intervention connects both views to externally observable responses. Future work should expand the prompt sets, incorporate automated behavioral evaluation alongside semantic review, and explore richer intervention strategies beyond simple top-five suppression. Nevertheless, the current results already show that MoE routing analysis can reveal meaningful internal structure related to safety behavior and can guide mechanism-oriented experiments in sparse language models.

\section*{Acknowledgment}
We would like to sincerely thank Dr. Yingfei Dong for his guidance, insightful ideas, and continuous support throughout this research project.

\balance
\end{document}